%% file: arxiv_main.tex
\documentclass[preprint,12pt]{elsarticle}

\usepackage{amssymb}
\usepackage{amsmath}
\usepackage{graphicx}
\usepackage{tabularx}
\usepackage{natbib}%
\usepackage{multirow}%
\usepackage{xcolor}%
\usepackage{textcomp}%
\usepackage{manyfoot}%
\usepackage{booktabs}%
\usepackage{graphbox}%
\usepackage{enumitem}%
\usepackage{wrapfig}%
\usepackage{lineno}%
\usepackage{subcaption}% [labelformat=simple]
\usepackage{changepage}
\usepackage{pdfpages}
\usepackage{xurl}
\usepackage{ragged2e}  % for '\RaggedRight' macro (allows hyphenation)
\newcolumntype{Y}{>{\RaggedRight\arraybackslash}X}

\usepackage[normalem]{ulem}  % For strikethrough (sout)

\begin{document}

\begin{frontmatter}



\title{Improving Equity in Health Modeling with GPT4-Turbo Generated Synthetic Data: A Comparative Study}

\author[label1]{Daniel Smolyak\corref{cor1}}
\ead{dsmolyak@umd.edu}
\cortext[cor1]{Corresponding author.}
\affiliation[label1]{organization={University of Maryland, College Park, Department of Computer Science},
            addressline={8125 Paint Branch Drive},
            city={College Park},
            state={MD},
            postcode={20742},
            country={USA}}

\author[label2,label5]{Arshana Welivita}
\ead{awelivi1@jhu.edu}
\affiliation[label2]{organization={Johns Hopkins University, Department of Computer Science},
            addressline={3400 North Charles Street},
            city={Baltimore},
            state={MD},
            postcode={21218},
            country={USA}}
\affiliation[label5]{organization={Center for Digital Health and Artificial Intelligence},
addressline={100 International Drive},
            city={Baltimore},
            state={MD},
            postcode={21202},
            country={USA}}

\author[label3]{Margrét V. Bjarnadóttir}
\ead{mbjarnad@umd.edu}
\affiliation[label3]{organization={University of Maryland, College Park, Robert H. Smith School of Business},
            addressline={4100 Mowatt Lane},
            city={College Park},
            state={MD},
            postcode={20740},
            country={USA}}

\author[label4,label5]{Ritu Agarwal}
\ead{ritu.agarwal@jhu.edu}
\affiliation[label4]{organization={Johns Hopkins University, Carey Business School},
            addressline={100 International Drive},
            city={Baltimore},
            state={MD},
            postcode={21202},
            country={USA}}

%% Abstract
\begin{abstract}
\textbf{Objective.} Demographic groups are often represented at different rates in medical datasets. These differences can create bias in machine learning algorithms, with higher levels of performance for better-represented groups. One promising solution to this problem is to generate synthetic data to mitigate potential adverse effects of non-representative data sets.

\textbf{Methods.} We build on recent advances in LLM-based synthetic data generation to create a pipeline where the synthetic data is generated separately for each demographic group. We conduct our study using MIMIC-IV and Framingham “Offspring and OMNI-1 Cohorts” datasets.  We prompt GPT4-Turbo to create group-specific data, providing training examples and the dataset context. An exploratory analysis is conducted to ascertain the quality of the generated data. We then evaluate the utility of the synthetic data for augmentation of a training dataset in a downstream machine learning task, focusing specifically on model performance metrics across groups. 

\textbf{Results.} The performance of GPT4-Turbo augmentation is generally superior but not always. In the majority of experiments our method outperforms standard modeling baselines, however, prompting GPT-4-Turbo to produce data specific to a group provides little to no additional benefit over a prompt that does not specify the group.

\textbf{Conclusion.} We developed a method for using LLMs out-of-the-box to synthesize group-specific data to address imbalances in demographic representation in medical datasets. As another ``tool in the toolbox'', this method can improve model fairness and thus health equity. More research is needed to understand the conditions under which LLM generated synthetic data is useful for non-representative medical data sets.

\end{abstract}

\begin{graphicalabstract}
\begin{figure}
    \includegraphics[scale=0.25]{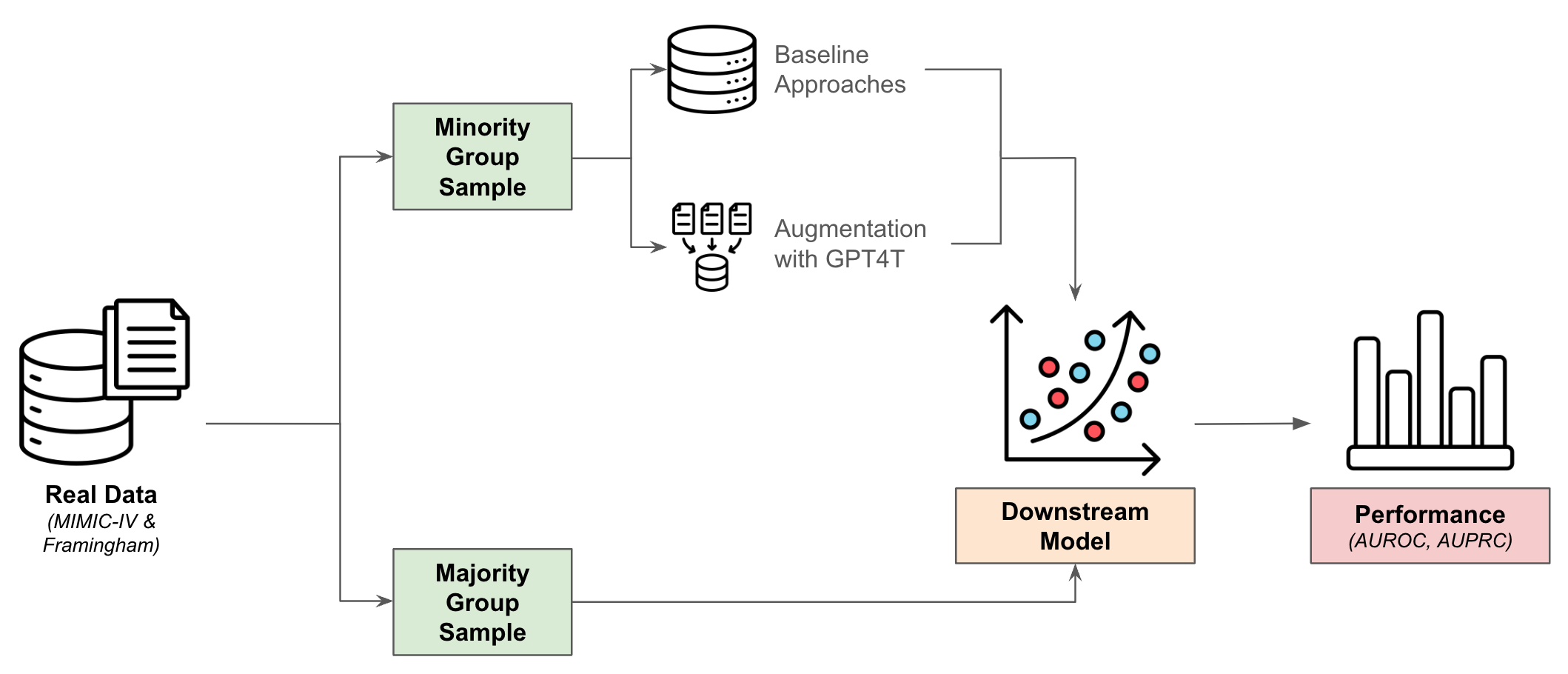}
    \label{fig:graph-abstract}
\end{figure}
\end{graphicalabstract}

% \begin{highlights}
% \end{highlights}

%% Keywords

\end{frontmatter}

\section{Introduction}

The growing interest in using machine learning (ML) for various healthcare prediction tasks has raised concerns about bias in training data that may adversely affect equity and predictive performance across sensitive populations \citep{zack_assessing_2024, huang_evaluation_2022, juhn_assessing_2022, park_comparison_2021}. One common source of performance variation in ML models across subpopulations is differences in the amount of data available on subgroups \citep{chen2018my}. Such data availability differences may be a result of a multitude of factors including i) subgroup sizes, ii) differences in subgroup disease prevalence, (by race, gender, or other demographic categories) \citep{movva_coarse_2023}, and iii) structural bias, which can lead to marginalized groups being systematically included in health datasets at lower rates \citep{nazha2019enrollment}. ML models developed using non-representative data sets amplify the risk of perpetuating and reinforcing health inequities \citep{kleinberg_racial_2022}, through worse model performance for underrepresented groups. 

The particular risk from modeling subgroups with differing data availability is that these models may prioritize predictive performance for larger groups. Models are typically optimized to return the highest average performance on the entire dataset, which disadvantages small groups, as their performance has less of an effect on the average. This is especially relevant in healthcare setting where different subgroups have different relationships between features and outcomes, such as gender differences in the signs of heart attacks \citep{devon2008symptoms}. Even when groups are weighted to contribute equally to the model objective, estimating a model from fewer samples results in decreased model performance (through increased statistical bias or variance \citep{chen2018my}). However, notions of fairness and equity in health settings require equal model performance across all groups. Thus, it is necessary to not only to monitor model performance across subgroups, but also to boost performance for smaller subgroups.

Synthetic data generation (SDG) has been proposed as a potential solution to the often limited size of healthcare data arising from privacy, financial considerations, or other constraints across a range of applications \citep{murtaza_synthetic_2023}. In the context of predictive healthcare modeling and the underperformance of models for specific subgroups, SDG offers a promising opportunity to augment datasets to improve these outcomes. Previously synthetic data has been used in medical image contexts, where the overall training dataset size is a limiting factor, and to address class imbalance, where certain outcomes have fewer examples \citep{chlap_review_2021}. Further, augmentation has been shown to improve fairness across demographic groups through either domain adaptation, where a generator learns to alter data from one group to match another group \citep{abhari2023mitigating}, or through conditional data generation, where a generator is trained to create data with certain prespecified attributes \citep{ktena2024generative}. For example, Ktena et al. \citep{ktena2024generative} develop image-generating diffusion models for three medical imaging contexts, which are able to generate data specific to a specified sex or racial group. These models are then used to augment a demographically-unbalanced training dataset to yield a more balanced synthetic dataset.

The recent release of powerful large language models (LLMs) has created new opportunities to generate synthetic health data with considerable ease and speed, beyond previous approaches (i.e., a shift from statistical models focused on recreating the data generation process, to more advanced deep learning methods). However, to date we have limited understanding of how synthetic data generated with LLMs can support predictive performance for small subgroups. The handful of studies that have leveraged LLMs for health data generation have focused largely on overall predictive model performance in downstream tasks, without specifically considering performance within subgroups. In this context, Borisov et al., \cite{borisov_language_2022} show that smaller LLMs (e.g. GPT2) can be fine-tuned on tabular data sets to produce related synthetic data and Seedat et al. \cite{seedat_curated_2024} study the ability of GPT4 to generate tabular data without fine-tuning, offering evidence that LLMs can generalize to generate data for subgroups not provided as examples. Kim et al. \cite{kim2024group} use LLMs (including GPT3.5, LLaMa2, and Mistral) to generate tabular data in settings with outcome class imbalance, improving performance for an underrepresented class in the binary setting. In contrast to these papers, our focus is on whether we can use LLMs to improve outcomes for underrepresented groups in downstream analytical tasks.

In this paper, we build upon the focus on tabular health data generation by LLMs of these previous studies. We pose the question: can SDG using LLMs improve subgroup predictive performance in tabular health data settings? Given the limited evidence to date, an understanding of the comparative value of LLMs for healthcare modeling is critical for determining their applicability in different contexts. In this work we compare subgroup-specific data augmentation with other available modeling approaches that are often deployed by healthcare modelers and data scientists to improve predictive performance for subgroups. %\removemvb{To investigate the potential of SDG with LLMs for subgroup performance enhancement,} 
We conduct a series of comparative analyses using the MIMIC-IV and Framingham Heart Study data sets. We use the race identification data within each dataset to define different subgroups and examine whether synthetic data generated with GPT4-Turbo (henceforth, GPT4T for brevity) can improve prediction performance for underrepresented groups over standard modeling approaches including group weighting, building separate models for each group, and SMOTE \citep{chawla2002smote}.

Our results are mixed. In some cases, GPT4T-based synthetic data generation improves performance for smaller groups. However, in other instances, baseline approaches perform equally well or better. We conclude that augmentation approaches utilizing foundation models represent a complementary tool for addressing underperformance of predictive models for underrepresented groups.

\include{statement_of_significance}

%%%%%%%%%%%%%%%%%%%%%%%%%%%%%%%%%%%%%%%%%%%%
%%%%%%%%%%%%%%%%%%%%%%%%%%%%%%%%%%%%%%%%%%%%%

\section{Methods}

\subsection{Data and Dependent Variables}

We study the effectiveness of synthetic health data generation for prediction tasks on two publicly available and widely used health datasets. Our data set choice is driven by two core considerations: the availability of comparable race identifiers and documented differences in predictive model performance across racial groups. Across both data sets, we utilize self-identified race as our group variable, both datasets include the same 4 racial/ethnic groups, White, Asian, Black (or African American), and Hispanic (or Latino/Latinx). We designate White adults as the majority group, as across both data sets, there are far more White adults than adults from minority racial groups.

The first data set, MIMIC-IV is a publicly available electronic health record repository from the Beth Israel Deaconess Medical Center \citep{johnson2020mimic}. Prior research using MIMIC-IV data has shown racial group differences in prediction performance on several tasks, including hospitalization at triage, critical outcomes (ICU admission or death), and 3-day readmission \citep{movva_coarse_2023}. Movva et al. \citep{movva_coarse_2023} provide a pre-processing and modeling pipeline that we adapt, which includes the aforementioned three outcomes, and predictors such as primary complaint, comorbidities, and physical measurements taken at triage. 

Our second data set is the Offspring (OS) and OMNI 1 Cohorts from the Framingham Heart Study (FHS), a longitudinal study of heart disease and its potential causes \citep{tsao2015cohort}; the former cohort was recruited from the children of the (largely White) original cohort and the OMNI 1 Cohort was recruited specifically to introduce greater racial diversity into FHS. Previous research has shown racial differences in the magnitude of associations between risk factors and cardiovascular outcomes; for example, the association between diabetes and cardiovascular events is twice as high for Hispanic adults as for White adults \citep{gijsberts2015race}. Differences in associations indicate that a predictive model built on training data composed mostly of a single racial group (e.g. White adults) will inevitably perform worse for other racial groups. We predict three outcomes: 10-year diagnosis of coronary heart disease (CHD), 10-year diagnosis of cardiovascular disease (CVD), and 30-year diagnosis of congestive heart failure (CHF). We use the same set of features as Beunza et al. \citep{beunza2019comparison}, who compare machine learning methods to predict coronary heart disease risk at 10 years using the FHS data.

\subsection{LLM-based Synthetic Data Generation}

We follow the prompt structure outlined by Seedat et al. \citep{seedat_curated_2024} to generate synthetic data from relatively few examples with GPT4T. The prompt includes three sections: 1) the context of the dataset and prediction task, 2) examples of data, and 3) instructions for data generation. We provide 20 randomly selected examples from the minority group of interest in the original dataset, resampling if necessary to ensure that each outcome variable has at least one positive example within these 20. The prompt is then sent to GPT4T which returns synthetic data samples. 

We also consider the importance of explicitly identifying the minority group of interest within the prompt. In the context of image generation, previous studies have shown that including a group identifier in the prompt diversifies the resulting synthetic data \citep{clemmer_precisedebias_2024}. We therefore investigate whether including references to specific groups within the prompt produces data more closely resembling that group and the impact of such prompt engineering on predictive performance. More specifically we adjust the prompt from “generate data for 10 patients”, to a prompt with a subgroup label: “generate data for 10 Asian patients”. To examine whether the downstream benefits of data augmentation derive from group specification during data generation or simply from additional synthetic data, we also compare to a “group generic” approach, where no group is referenced in the prompt, and examples are sampled from the entire original dataset at random. Example prompts are included in the Appendix \ref{app:prompt} in the Supplementary Materials.

Prior to assessing the utility of the generated synthetic data on a downstream prediction task, we directly compare the original and synthetic data through exploratory analyses. First, we visually examine the correlation structure among key features in the original vs synthetic data.
We then investigate GPT4T’s capacity for generating synthetic data different from the original data, measured by the L1-distance between each synthetic data sample and the closest real data sample. Lastly, we evaluate whether our generated synthetic data is more similar to its intended minority group versus the majority group. To do so, we train a random forest model to predict race in the real datasets, and compare the predicted probabilities on the synthetic data to the predicted probabilities of hold-out sets of the real data for both groups.

\subsection{Subgroups}

For each of the predictive tasks, we designate White patients as the majority group and every other race/ethnicity (Asian, Black, and Hispanic) as minority groups. We accordingly sample a much smaller proportion of patient records from the minority group compared to the majority group. We then run separate experiments with each of the other 3 minority racial/ethnic groups available in both datasets as the minority groups. For each experiment we randomly selected 100 individuals from each minority group and 1000 individuals from the majority group. We provide sensitivity analysis of the minority group size in the Supplementary Materials (\ref{app:min-size}).

\subsection{Machine Learning Modeling}

\begin{figure}
\centering
\includegraphics[scale=0.3]{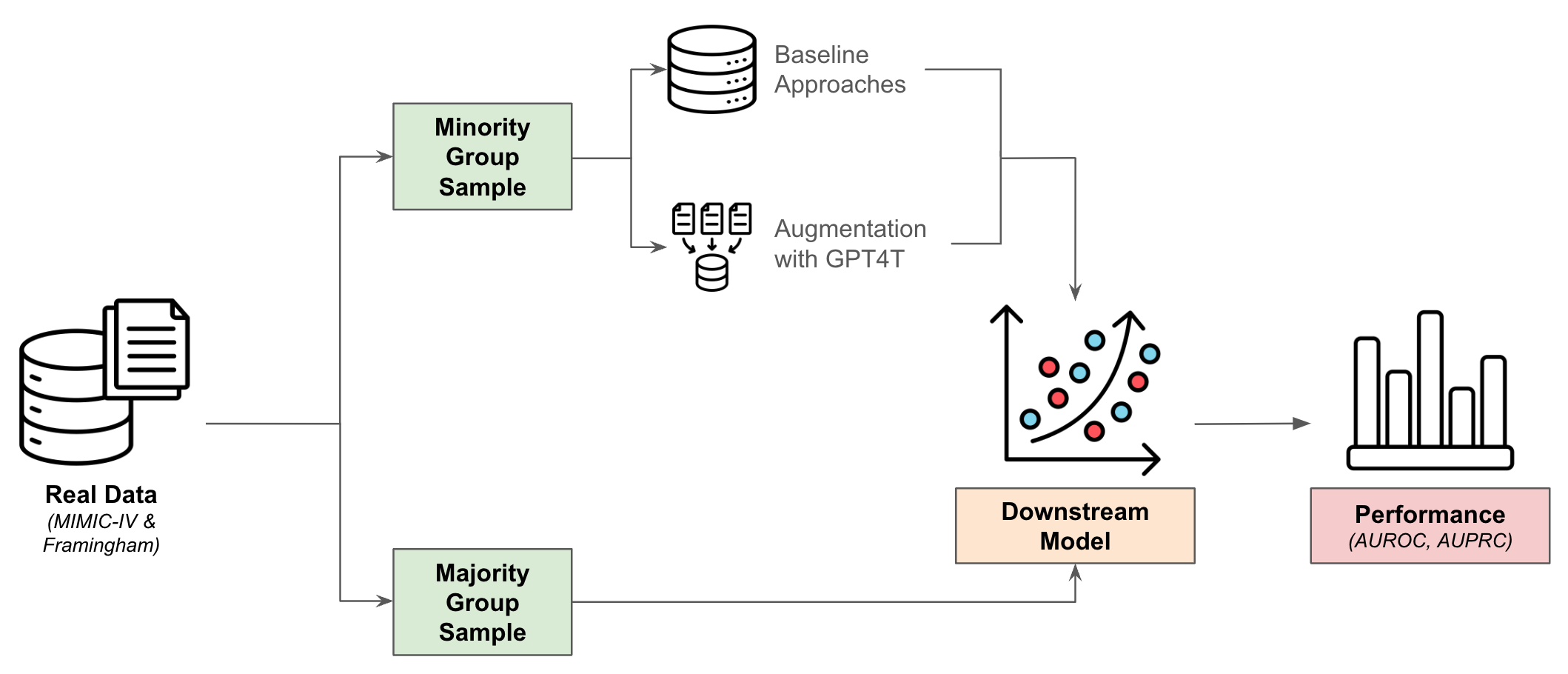}
\caption{Pipeline overview: first the data from the designated majority group and minority group are sampled,  
next the minority group data is either augmented with GPT4T-generated synthetic data (our approach), one of the baseline approaches is applied to the data, or no change is applied. The groups are then combined and used to train a logistic regression model.
The performance of the model is evaluated for each group on the held-out testing data.}
\label{fig:llm-pipeline}
\end{figure}

For each outcome and each minority group, we run a logistic regression model with a binary group indicator on the combined majority and minority data (without data augmentation) and evaluate the model’s performance for each group. We then compare this baseline to several other approaches to boosting small group performance, including our synthetic data generation and augmentation approach.  
In contexts where modelers are interested in increasing the performance of a smaller group, there are many approaches across a range of increasing complexity. In the absence of synthetic data, one might use separate models for each group \citep{ustun2019fairness} or upweight smaller groups during model estimation \citep{bastani2021predicting,dondelinger2020joint}. We also included SMOTE \citep{chawla2002smote}, an interpolation approach for generating synthetic data, where synthetic samples are identified as data points “between” existing samples. We use the sampling strategy where the minority group is up-sampled to the size of the majority group. The full pipeline is shown in Figure \ref{fig:llm-pipeline}.

\subsection{Performance Metrics}
We compare the predictive performance for each outcome using the area under the receiver operator curve (AUROC) and area under the precision-recall curve (AUPRC). These metrics are particularly useful in our settings because of the imbalance of positive versus negative outcomes (e.g. given the relatively low incidence of CHD, there are many fewer participants who experience coronary heart disease than those who do not), and the relatively higher importance of correctly identifying positive outcomes versus negative outcomes (e.g. identifying patients at risk of CHD)\citep{chen2021ethical}. Performance on the predictive task is averaged over 25 experiments for each setting, with a different training sample in each iteration. We evaluate these performance metrics on a hold-out testing
set that includes all remaining samples after removing the samples provided to GPT4T and the samples used to train the downstream model (subgroup sizes for each dataset are available in the Supplementary Materials in Table \ref{tab:data-size}). We do not include results for the CHF outcome for Asian participants in the Framingham datasets due to an insufficient number of CHF cases (3 out of 111 total Asian participants).

\section{Results}
\label{sec:results}

Prior to evaluating the downstream utility of the GPT4T-generated synthetic data, we seek to understand how well the synthetic data replicates the correlation structure of the real data, whether it simply copies the real data, and the ability of GPT4T to generate data more similar to the minority group than the majority group. 
We focus largely on Black patients in MIMIC-IV and Hispanic participants in Framingham as a case study.
While the downstream utility is the eventual goal, this exploratory analysis further illuminates the strengths and weaknesses of the generation method.

\subsection{Quality of Synthetic Data}

\begin{figure*}
    \centering
     \includegraphics[scale=0.4]{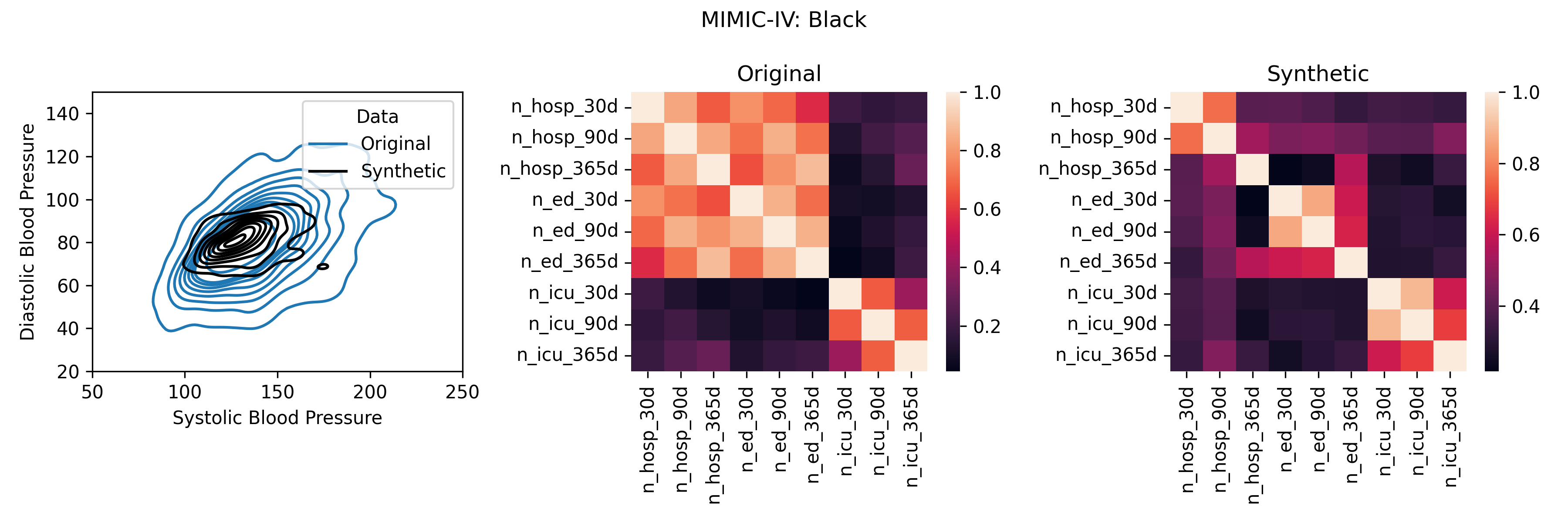}
     \includegraphics[scale=0.4]{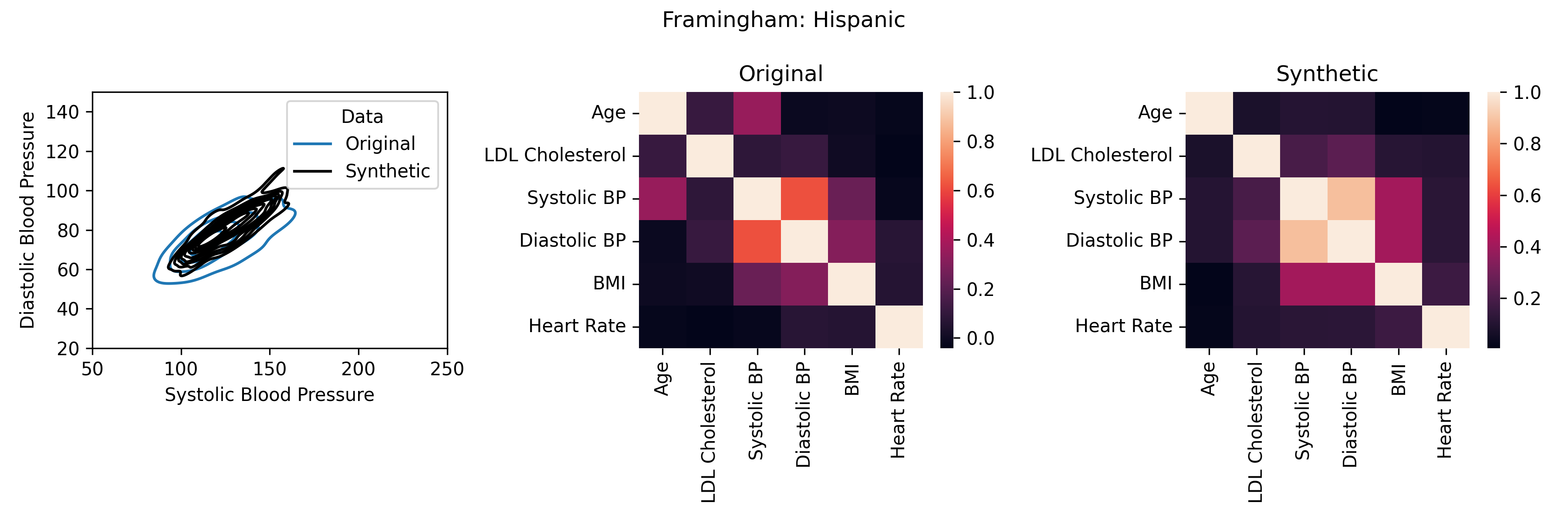}
     \caption{(left) Kernel density estimation for the joint distribution of diastolic and systolic blood pressure in (left top) the Black-tailored GPT4T-generated data versus the real MIMIC-IV data for Black patients and (left bottom) Hispanic-tailored GPT4T-generated data versus the real Framingham data for Hispanic participants. (top center) The correlations plot for previous hospital, emergency department, and ICU admissions in the original MIMIC-IV data and (top right) the GPT4T-generated data. (bottom center) The correlations plot for several heart health related variables in the original Framingham data and (bottom right) the GPT4T-generated data. 
     }
     \label{fig:llm-corr}
\end{figure*}

We first conducted an exploratory analysis, %\removemvb{of the data generated by GPT4T. Figure \ref{fig:llm-corr}}
comparing the correlation structure in the original data and the generated data. Figure \ref{fig:llm-corr} highlights that GPT4T replicates some of the correlation structure of the original data. GPT4T preserves the modest correlation between systolic and diastolic blood pressure in MIMIC-IV (Fig. \ref{fig:llm-corr} top left), but the range of synthetic blood pressure values coalesces more closely around the average. Meanwhile, GPT4T better overlaps the range for blood pressure values for the Framingham data (Fig. \ref{fig:llm-corr} bottom left). Similarly, GPT4T replicates many but not all of the correlations between the number of past hospitalizations, emergency department visits and ICU admissions in the past 30, 90, and 365 days in the MIMIC-IV data (Fig. \ref{fig:llm-corr} top center and right). It successfully produces some of the strong correlations within each type of visit, however, between visit types, specifically between past hospitalizations and past emergency department visits, the correlations produced by GPT4T are not as strong or missing compared to the original data. For the Framingham data GPT4T again mostly replicates the correlation structure (Fig. \ref{fig:llm-corr} bottom center and right). The notable differences are that GPT4T generates data with stronger correlations between both blood pressure values and between those values and BMI, compared to the real data. Additionally, moderate correlation between age and systolic blood pressure in the real data is not captured in the synthetic data. Thus, the generated synthetic data is able to capture many key relationships (or lack thereof) between features, but does not represent a perfect replication of the overall correlation structure. 

\begin{figure*}[]
    \centering
    \includegraphics[scale=0.4]{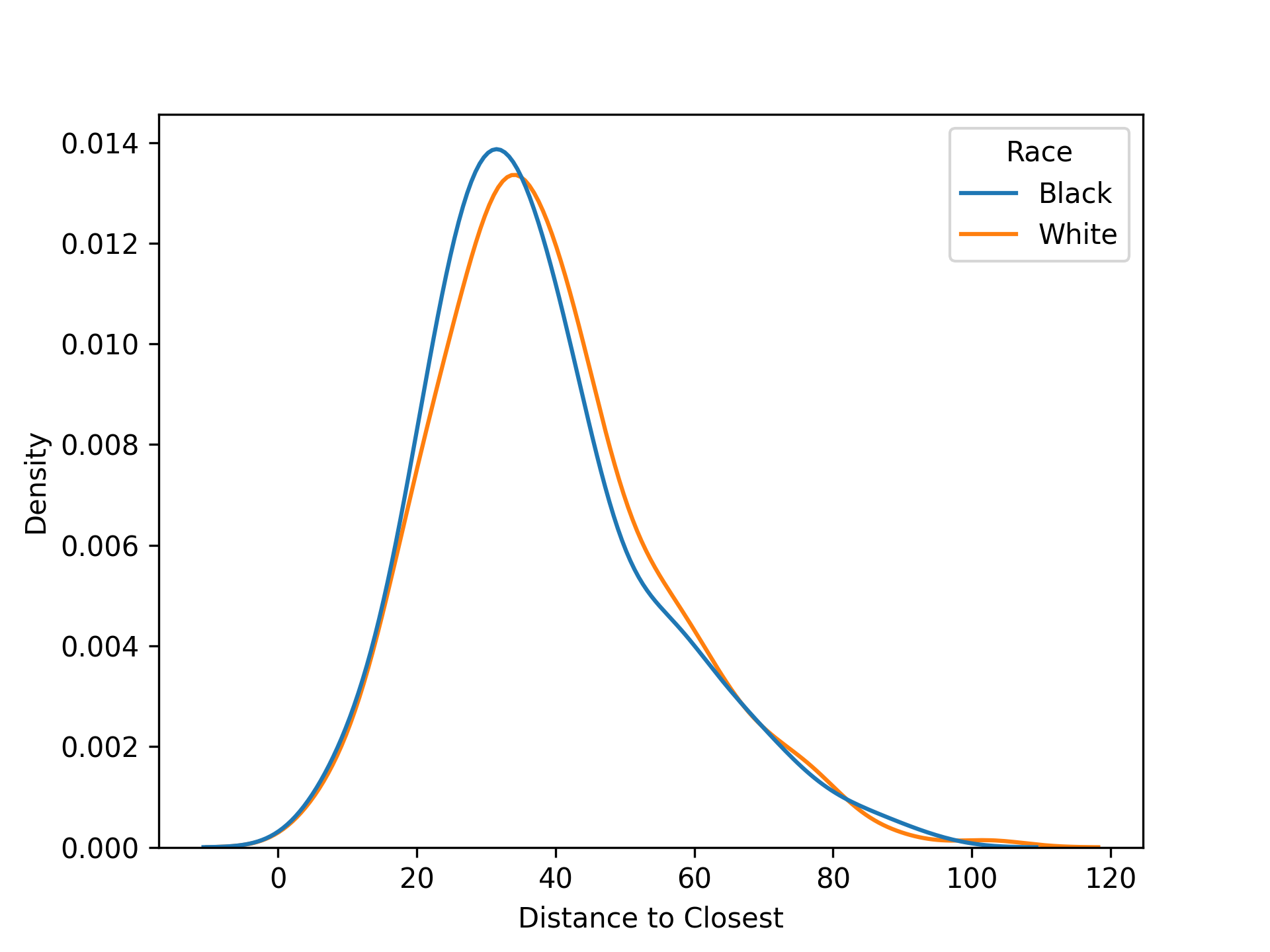}
    \includegraphics[scale=0.4]{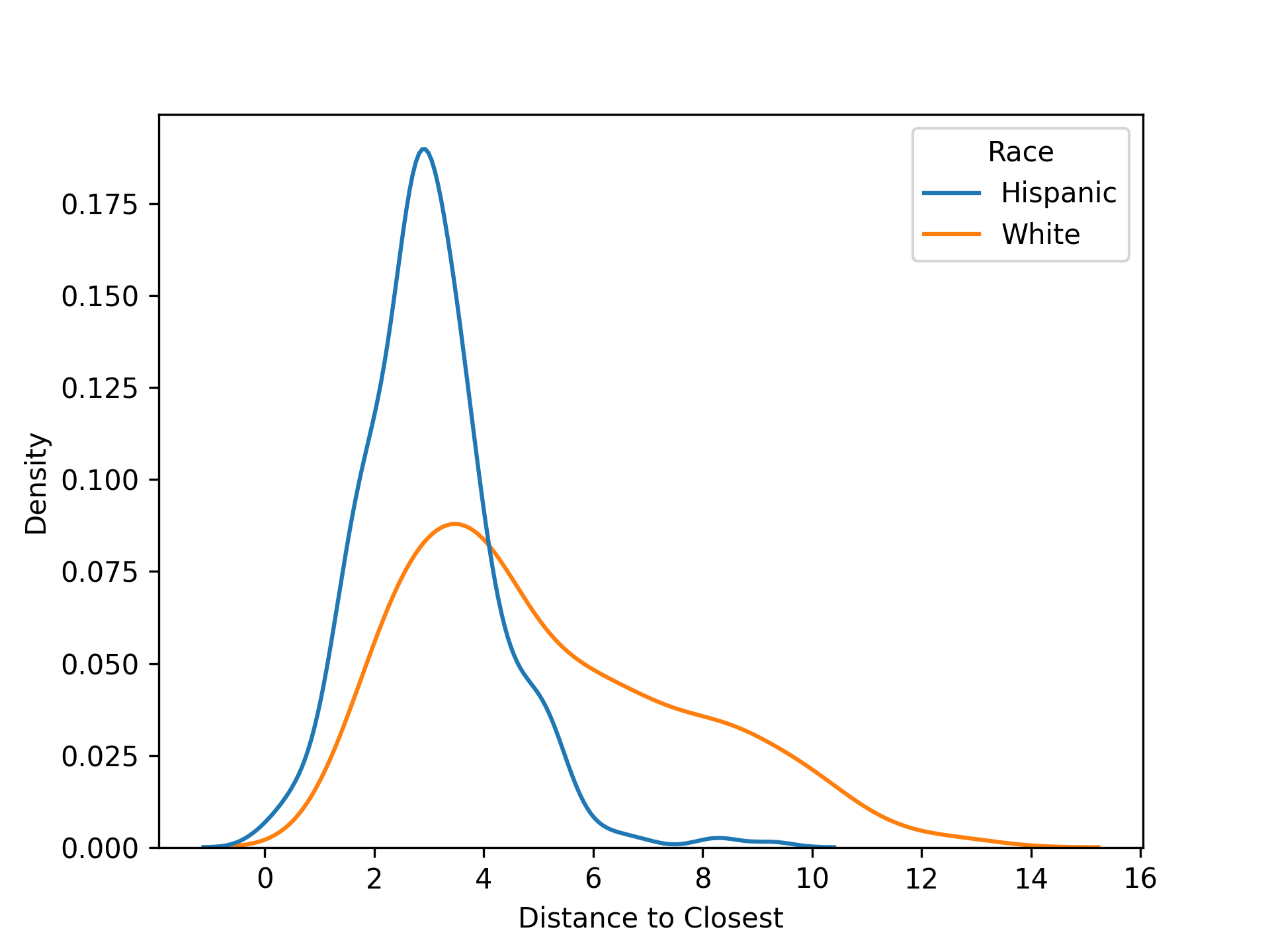}
     \caption{(left) MIMIC-IV: Distribution of the distances between each GPT4T generated data point (with a Black-specific prompt) and the nearest sample of real Black patients and real White patients. (right) Framingham: Distribution of the distances between each GPT4T generated data point (with a Hispanic-specific prompt) and the nearest sample of real Hispanic patients and real White patients.}
     \label{fig:llm-dist}
\end{figure*}

We next summarize the distance between the generated data and the real world data. In Figure \ref{fig:llm-dist} we summarize the L1-distance between each GPT4T sample and its closest real data sample for both datasets for the same subpopulations as before. For both datasets we observe that GPT4T generates “new” samples that are not copies of existing data, as very few synthetic samples have a distance to the nearest real data sample close to 0. For the Framingham data, GPT4T generates data that is closer to Hispanic participants than White participants. However, for MIMIC-IV, the GPT4T data does not appear to be more similar to either racial group (Black or White).

\begin{figure}
    \centering
    \includegraphics[scale=0.5]{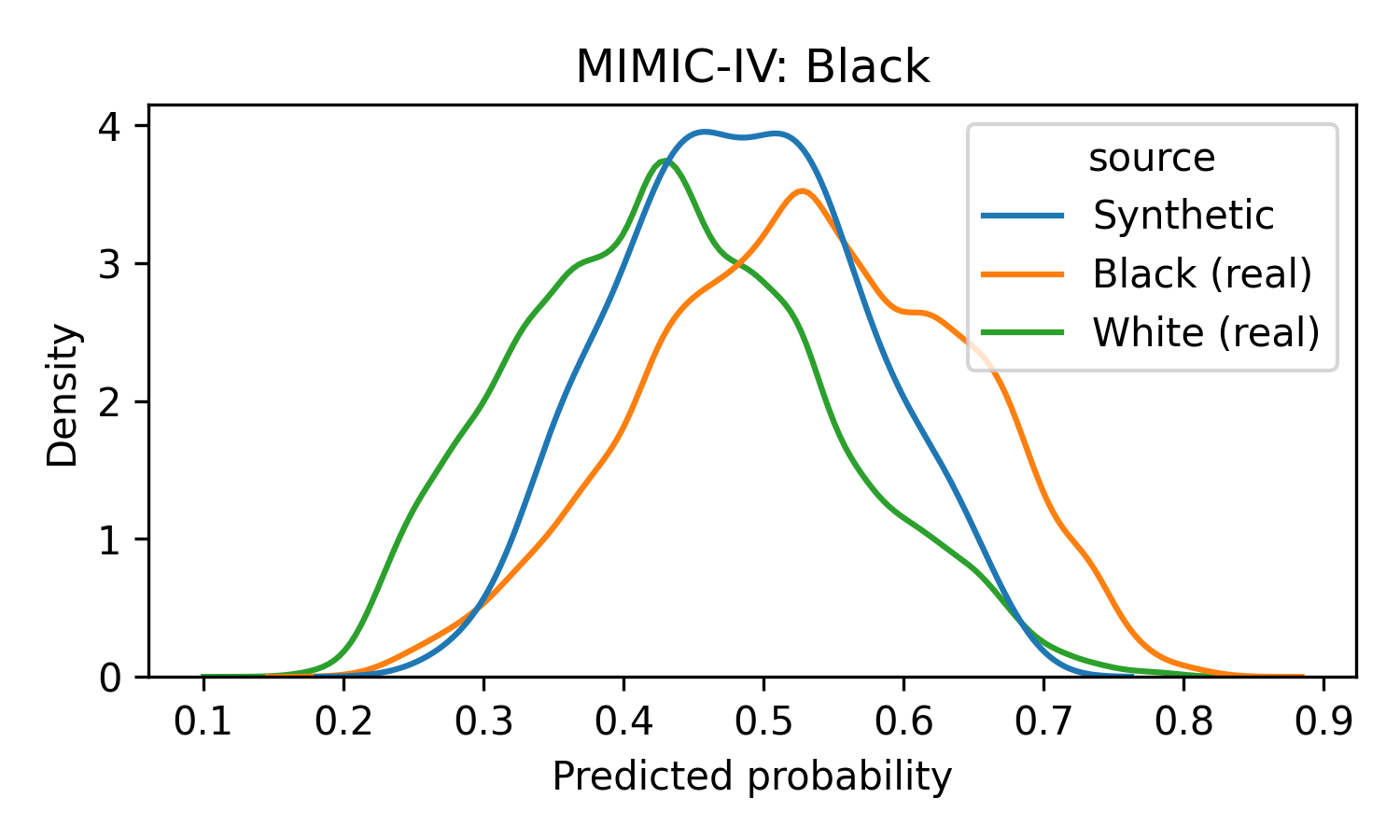}
    \includegraphics[scale=0.5]{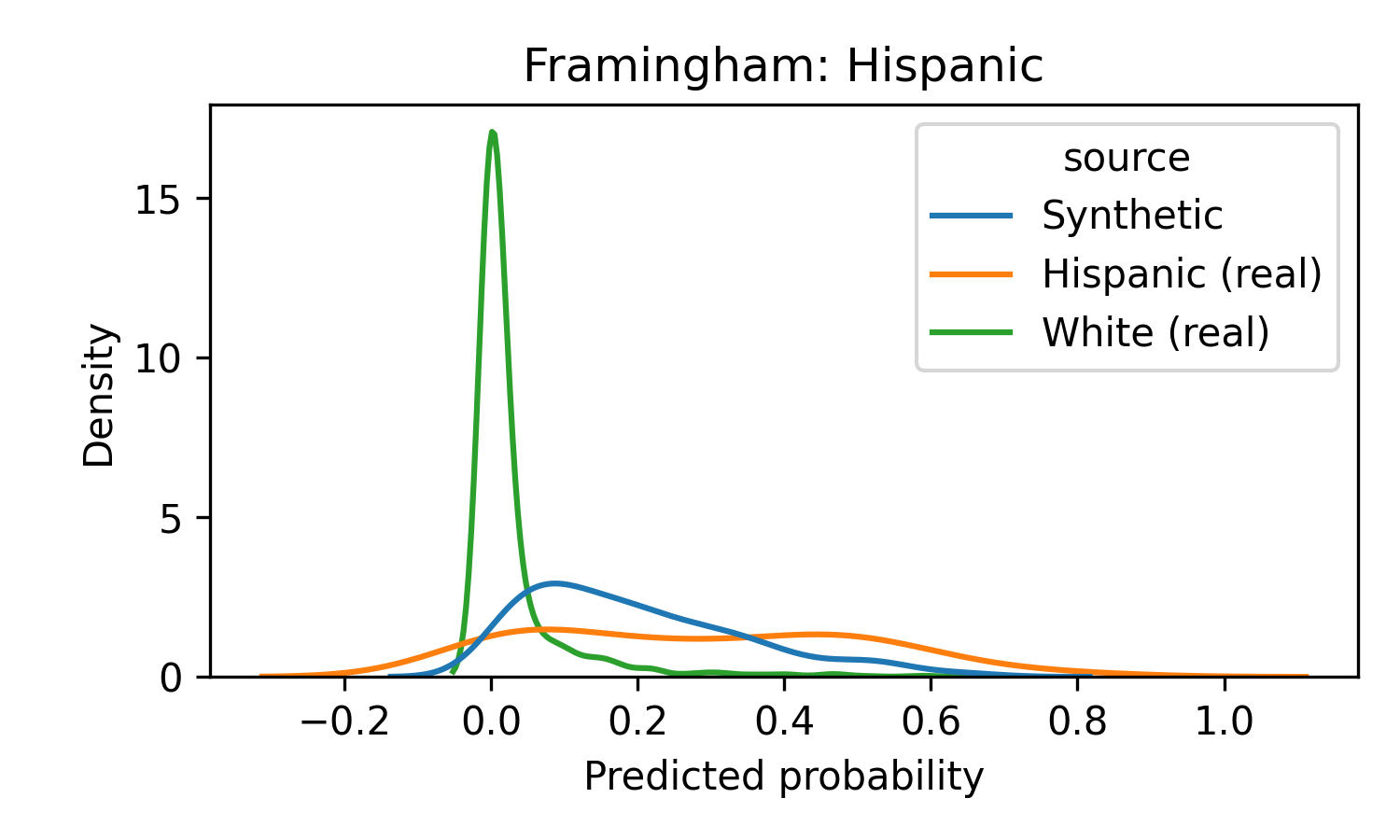}
    \caption{Density of predicted probabilities for belonging to the minority racial group for the synthetic data (blue), each minority group (orange), and the majority group (green) in the MIMIC-IV data (left) and the Framingham data (right).}
    \label{fig:llm-rf-dist}
\end{figure}

In Figure \ref{fig:llm-rf-dist} we further examine whether the data generated is more similar to the intended minority group. We train a random forest model to predict each individual's race (focusing on just Black vs White patients in MIMIC-IV and Hispanic vs White participants in Framingham). We then predict the probability of each individual belonging to the minority group on an out-of-sample set of the original data, and we predict the same probabilities for the synthetic data generated for that minority group. Plotting the distribution of these predicted probabilities, we would expect the majority group predictions to cluster towards 0 and the minority group predictions to cluster towards 1 (for a model that is able to easily distinguish racial groups).

In visualizing the density of predicted racial group probabilities for each data source, we find that the GPT4T-generated synthetic data falls between minority and majority groups for both the Hispanic patients in Framingham and the Black patients in MIMIC-IV. In MIMIC-IV, the Black and White patients already have a large overlap in predicted probabilities, and the synthetic data greatly overlaps with both. In Framingham, the White participants are actually clustered towards 0 while the Hispanic participants are evenly spread over the distribution from 0 to 1. The synthetic data skews slightly towards 0, but is much more evenly distributed than the White participants.

\subsection{Predictive Performance}

Table \ref{tab:llm-auroc} summarizes the results of GPT4T-generated synthetic data augmentation compared with our baselines, for the AUROC metrics (results for the AUPRC metric are summarized in Table \ref{tab:llm-auprc} in the Supplementary Materials, as the insights and conclusions are very similar.) From Table \ref{tab:llm-auroc} we see that in a majority of cases, 13 out of 17 for AUROC (and 9 out of 17 for AUPRC), models using LLM augmented data perform better than baselines, including for all groups and outcomes in MIMIC-IV and all outcomes in Framingham for Hispanic patients (for AUROC). The improvement over the baseline ranges up to a 0.1121 improvement in AUROC (17\%) for Hispanic participants in Framingham for the CHD outcome. The largest improvement in the MIMIC-IV dataset is a 0.0255 improvement in AUROC (4.1\%) for Black participants for the ed\_revisit\_3d outcome. However, within these cases, prompting GPT4T to generate data for a specific racial group provides limited additional benefit in improving prediction. In 7 out of 13 cases for AUROC, the prompt without a specific racial group outperforms the race tailored data (5 out of 9 for AUPRC). Additionally in most cases (14 out of 17), AUROC differences are less than 0.01 between the generic and tailored prompts. The improvements in AUROC from the baseline to either of the LLM-based SDG methods are also in some cases small. 
Examining the other standard approaches, SMOTE was almost never the highest performing method for either metric (the only exception is AUPRC for Black participants in Framingham for the CHD outcome). Upweighting and separate models were the highest performing method for one to three cases for each metric. To summarize, in the majority of experiments GPT4T generation outperformed the baseline and the other standard approaches, but the improvement from group-specific data generation was often limited. 

\begin{table}[!ht]
\begin{adjustwidth}{-1.5cm}{-1.5cm}
    \centering
    \scriptsize
    \begin{tabularx}{\linewidth}{lllcccccc} % |l|l|l|c|c|c|c||c|c|
        \toprule
        & & & &\multicolumn{3}{c}{Standard Approaches} & \multicolumn{2}{c}{GPT4T Generation} \\ \cmidrule(lr){5-7} \cmidrule(lr){8-9}
        \textbf{Dataset} & \textbf{Race} & \textbf{Outcome} & \textbf{Baseline} & \textbf{Upweighted} & \textbf{Separate} & \textbf{SMOTE} & \textbf{Group} & \textbf{Generic} \\ \cline{1-9}
        \multirow[c]{9}{*}{MIMIC-IV} & \multirow[c]{3}{*}{Asian} & critical & 0.8744 & 0.8638 & 0.8733 & 0.8661 & 0.8777 & \bfseries 0.8782 \\
 &  & ed\_revisit\_3d & 0.5681 & 0.5498 & 0.5624 & 0.5501 & \bfseries 0.5693 & 0.5646 \\
 &  & hospitalization & 0.8130 & 0.8100 & 0.8132 & 0.8092 & 0.8132 & \bfseries 0.8139 \\ \cline{2-9}
 & \multirow[c]{3}{*}{Black} & critical & 0.8563 & 0.8524 & 0.8551 & 0.8481 & \bfseries 0.8609 & 0.8576 \\
 &  & ed\_revisit\_3d & 0.6250 & 0.5925 & 0.6125 & 0.5871 & \bfseries 0.6505 & 0.6430 \\
 &  & hospitalization & 0.7990 & 0.7981 & 0.7975 & 0.7966 & \bfseries 0.8024 & 0.8017 \\ \cline{2-9}
 & \multirow[c]{3}{*}{Hispanic} & critical & 0.8486 & 0.8419 & 0.8464 & 0.8373 & 0.8534 & \bfseries 0.8554 \\
 &  & ed\_revisit\_3d & 0.5858 & 0.5684 & 0.5844 & 0.5674 & 0.6075 & \bfseries 0.6104 \\
 &  & hospitalization & 0.7915 & 0.7895 & 0.7893 & 0.7870 & 0.7915 & \bfseries 0.7925 \\
 \cline{1-9}
        \multirow[c]{8}{*}{Framingham} & \multirow[c]{2}{*}{Asian} & CHD & 0.6672 & \bfseries 0.6971 & 0.6461 & 0.6489 & 0.6867 & 0.6396 \\ 
 &  & CVD & 0.7292 & 0.7257 & 0.7600 & 0.7167 & \bfseries 0.7770 & 0.7728 \\ \cline{2-9}
 & \multirow[c]{3}{*}{Black} & CHD & \bfseries 0.6028 & 0.5437 & 0.5928 & 0.5959 & 0.5791 & 0.5869 \\
 &  & CVD & 0.6796 & \bfseries 0.7003 & 0.6075 & 0.6883 & 0.6172 & 0.6434 \\
 &  & CHF & 0.6531 & 0.6211 & \bfseries 0.6783 & 0.6488 & 0.6163 & 0.6389 \\ \cline{2-9}
 & \multirow[c]{3}{*}{Hispanic} & CHD & 0.6778 & 0.6842 & 0.6131 & 0.6721 & \bfseries 0.7903 & 0.7899 \\
 &  & CVD & 0.6874 & 0.6755 & 0.6563 & 0.6679 & 0.7312 & \bfseries 0.7382 \\
 &  & CHF & 0.8144 & 0.8163 & 0.6661 & 0.7988 & 0.8685 & \bfseries 0.8831 \\ \bottomrule
    \end{tabularx}
    \caption{\textbf{AUROC} scores for each dataset, minority racial group, outcome variable, and method for augmenting small groups. Abbreviations for outcomes are critical: Critical event (ICU or death), ed\_revisit\_3d: 3-day emergency department readmission, hospitalization: hospital admission, CHD: coronary heart disease, CVD: cardiovascular disease, CHF: coronary heart failure.
    }
    \label{tab:llm-auroc}
\end{adjustwidth}
\end{table}

\section{Discussion}

GPT4T-based synthetic health data generation presents opportunities for augmentation of datasets, particularly when focusing on underrepresented populations. We found that while GPT4T generated data often improves predictive performance for these groups compared to both the original data and common baselines, the improvements are typically small. This indicates that out-of-the-box GPT4T generation can be considered “another tool in the toolkit” of available strategies for boosting small group performance, rather than a consistently superior method. 

Our analysis of race-specific prompts also yielded mixed results, in some cases race-specific prompts outperformed generic prompts but not consistently so. Unlike the context of diverse image generation \citep{clemmer_precisedebias_2024}, the differences between groups and their predictive performance in the tabular setting are more complex. As investigated in Movva et al. \citep{movva_coarse_2023}, group differences in MIMIC-IV were attributable to a variety of factors, including differing distributions/frequencies of feature and outcome variables, and varying relationships between features and outcomes. Similar differences between racial and ethnic groups in the heart disease context have been reported as well \citep{gijsberts2015race}. Given our mixed results, it is reasonable to conclude that GPT4T was unable to produce data that meaningfully captured all nuances of these differing distributions and relationships found in a given minority group but not the majority group, however increasing minority data with synthetic data in more than half the cases improved predictions for the minority group.

LLMs present unique risks in the area of SDG. Past work has shown that GPT4 replicates societal bias in generation of clinical vignettes \citep{zack_assessing_2024}, suggesting that such biases may manifest in tabular health data generation as well. We similarly found that group-unaware GPT4T-based data generation (“GPT4T Generic”) risks exacerbating group differences in performance. For example, GPT4T Generic increased AUROC by at least 8\% across all outcomes for Hispanic individuals in the Framingham dataset, while performance fell across outcomes for Black individuals. This effect prompts further concern when considering that AUROC was already higher for Hispanic individuals compared to Black individuals prior to augmentation (“Original”), highlighting the importance of considering group membership throughout the machine learning pipeline, and even adjusting the modeling approach to each group. While one might use GPT4T augmentation across outcomes for Hispanic participants in the Framingham dataset, it would be advisable to use one of the standard approaches for Black participants (e.g. upweighting for the CVD outcome). 
Our findings are consistent with what has been shown in many other contexts in the fair machine learning literature, group-unawareness throughout the modeling pipeline can lead to unfair performance differences \citep{chen2021ethical,mehrabi2021survey}.

There are several directions to consider for improving our proposed approach for group-specific data generation, that may be fruitful future research directions. The prompt could more explicitly refer to the expected data differences in the group of focus, leveraging prior research on chain-of-thought reasoning \citep{chung2024scaling}. Health-specific models, such as MedPaLM-2 \citep{singhal2023towards}, may have a better contextual understanding of data differences from the medical research in their training corpuses. Similarly, a retrieval augmented generation approach \citep{chen2024benchmarking} can be applied, where the LLM is required to reference a specific medical text that investigates the group differences and reflects that information in the generation. Additionally, Kim et al. \citep{kim2024group} provide a novel prompt structure, where positive and negative outcome classes are presented in separate sections within the prompt to help the LLM distinguish between the classes. A similar approach can be experimented with in our setting, providing examples from both a minority and majority subgroup within the same prompt. The focus on outcome classes can also be applied to the subgroup setting, exploring whether balanced generation of outcome classes within each subgroup also helps improve subgroup performance.

\section{Conclusion}

We outlined an easy-to-implement GPT4T-based synthetic data generation method that requires few real data examples and produces realistic samples. This method can be successfully used to augment smaller, underrepresented groups in health datasets.  However, our experiments showed that the prediction improvement is not consistent, in some cases performing on par with or worse compared to other baselines. Thus, our suggested approach should be considered as an additional tool in a modeler's toolkit, which may (or may not) improve predictive performance of small groups. However as our results indicate, the improvement in some cases can be significant. %\removemvb{caution is required when considering adding this generation method as a tool to support modeling and data preprocessing systems}. 
More research is needed to fully understand the conditions under which out-of-the-box foundation LLM models are useful across a wide range of SDG tasks for addressing health equity.

\bibliographystyle{elsarticle-num}
\bibliography{bibliography}

\appendix

\setcounter{table}{0}
\renewcommand{\thetable}{S\arabic{table}}

\setcounter{figure}{0}
\renewcommand{\thefigure}{S\arabic{figure}}

\section{Dataset Characteristics}
\label{app:data}

We provide the characteristics of each dataset in following tables. Table \ref{tab:data-size} includes the size of each racial subgroup, and Table \ref{tab:outcome-prev} include the percentage of positive (true) cases for each outcome for each subgroup. Both the subgroup sizes and prevalances differ vastly between subgroups.

\begin{table}[p]
\centering
\begin{tabular}{lcccc}
    \toprule
     & \textbf{Asian} & \textbf{Black} & \textbf{Hispanic} & \textbf{White} \\ \cline{1-5}
    \textbf{MIMIC-IV} & 18,321 & 92,168 & 35,205 & 244,093 \\
    \textbf{Framingham} & 111 & 170 & 215 & 2927 \\ \bottomrule
\end{tabular}
\caption{Number of individuals within each dataset, divided by racial group.}
\label{tab:data-size}
\end{table}

\begin{table}[]
    \centering
\begin{tabular}{lcccccc}
\toprule
& \multicolumn{3}{c}{Framingham} & \multicolumn{3}{c}{MIMIC-IV} \\ \cmidrule(lr){2-4} \cmidrule(lr){5-7}
 & \textbf{CHD} & \textbf{CHF} & \textbf{CVD} & \textbf{critical} & \textbf{ed\_revisit\_3d} & \textbf{hospitalization} \\
\midrule
\textbf{Asian} & 11.7 & 2.7 & 14.4 & 5.0 & 3.1 & 38.9 \\
\textbf{Black} & 10.6 & 7.1 & 20.6 & 3.9 & 4.7 & 39.3 \\
\textbf{Hispanic} & 7.9 & 4.2 & 11.6 & 2.9 & 3.9 & 34.9 \\
\textbf{White} & 2.0 & 1.5 & 2.7 & 6.7 & 3.3 & 53.3 \\
\bottomrule
\end{tabular}
\caption{Percentage of each subgroup for which the outcome is true/positive for each dataset.}
\label{tab:outcome-prev}
\end{table}

\section{AUPRC Results}
\label{app1}

Table \ref{tab:llm-auprc} summarizes the results of GPT4T-generated synthetic data augmentation compared with other standard approaches, for the AUPRC metric. As discussed in the main text, in a majority of cases (9 out of 17), models using LLM augmented data perform better than the baseline. The improvement over the baseline ranges up to a 0.168 improvement in AUPRC (62.7\%) for Hispanic participants in Framingham for the CHD outcome. The largest improvement in the MIMIC-IV dataset is a 0.0206 improvement in AUPRC (21.7\%) for Hispanic participants for the ed\_revisit\_3d outcome.

\begin{table}[ht]
\begin{adjustwidth}{-1.5cm}{-1.5cm}
    \centering
    \scriptsize
    \begin{tabularx}{\linewidth}{lllcccccc} % |l|l|l|c|c|c|c||c|c|
        \toprule
        & & & \multicolumn{4}{c}{Standard Approaches} & \multicolumn{2}{c}{GPT4T Generation} \\ \cmidrule(lr){4-7} \cmidrule(lr){8-9}
        \textbf{Dataset} & \textbf{Race} & \textbf{Outcome} & \textbf{Baseline} & \textbf{Upweighted} & \textbf{Separate} & \textbf{SMOTE} & \textbf{Group} & \textbf{Generic} \\ \cline{1-9}
        \multirow[c]{9}{*}{MIMIC-IV} & \multirow[c]{3}{*}{Asian} & critical & \bfseries 0.3457 & 0.3193 & 0.3440 & 0.3200 & 0.3385 & 0.3343 \\
 &  & ed\_revisit\_3d & 0.0401 & 0.0377 & 0.0395 & 0.0378 & 0.0395 & \bfseries 0.0403 \\
 &  & hospitalization & 0.7345 & 0.7297 & 0.7343 & 0.7284 & \bfseries 0.7356 & 0.7356 \\ \cline{2-9}
 & \multirow[c]{3}{*}{Black} & critical & 0.2685 & 0.2543 & 0.2663 & 0.2455 & \bfseries 0.2688 & 0.2546 \\
 &  & ed\_revisit\_3d & 0.1403 & 0.1246 & 0.1372 & 0.1228 & 0.1534 & \bfseries 0.1570 \\
 &  & hospitalization & \bfseries 0.7100 & 0.7056 & 0.7068 & 0.7029 & 0.7093 & 0.7096 \\ \cline{2-9}
 & \multirow[c]{3}{*}{Hispanic} & critical & 0.2037 & 0.1899 & \bfseries 0.2040 & 0.1882 & 0.1931 & 0.1898 \\
 &  & ed\_revisit\_3d & 0.0950 & 0.0830 & 0.0958 & 0.0805 & 0.1073 & \bfseries 0.1156 \\
 &  & hospitalization & \bfseries 0.6681 & 0.6638 & 0.6654 & 0.6598 & 0.6641 & 0.6676 \\ \cline{1-9}
\multirow[c]{8}{*}{Framingham} & \multirow[c]{2}{*}{Asian} & CHD & 0.4671 & 0.5479 & 0.4699 & 0.5131 & \bfseries 0.5738 & 0.4473 \\
 &  & CVD & 0.5846 & 0.5842 & \bfseries 0.6537 & 0.5477 & 0.6536 & 0.6362 \\ \cline{2-9}
 & \multirow[c]{3}{*}{Black} & CHD & 0.2069 & 0.1939 & 0.1964 & \bfseries 0.2158 & 0.2109 & 0.1818 \\
 &  & CVD & 0.4423 & \bfseries 0.4517 & 0.3711 & 0.4273 & 0.3896 & 0.3837 \\
 &  & CHF & 0.2246 & 0.1390 & \bfseries 0.2492 & 0.1981 & 0.2447 & 0.2141 \\ \cline{2-9}
 & \multirow[c]{3}{*}{Hispanic} & CHD & 0.2678 & 0.2921 & 0.2021 & 0.2404 & \bfseries 0.4358 & 0.3784 \\
 &  & CVD & 0.3030 & 0.2966 & 0.3098 & 0.2874 & \bfseries 0.4002 & 0.3854 \\
 &  & CHF & 0.2874 & 0.2722 & 0.2428 & 0.2673 & 0.3166 & \bfseries 0.4235 \\ \bottomrule
    \end{tabularx}
    \caption{\textbf{AUPRC} scores for each dataset, minority racial group, outcome variable, and method for augmenting small groups. Outcome abbreviations, critical: Critical event (ICU or death), ed\_revisit\_3d: 3-day emergency department readmission, hospitalization: hospital admission, CHD: coronary heart disease, CVD: cardiovascular disease, CHF: coronary heart failure.}
    \label{tab:llm-auprc}
\end{adjustwidth}
\end{table}

\section{GPT4T Prompt Structure}
\label{app:prompt}

We provide in Supplementary Table \ref{tab:prompt} an example prompt, illustrating the structure described in the main text. We first specify the role of the model as a synthetic data generator. Then the relevant context for the generation task is added, ``hospital admissions and readmission'' for the MIMIC-IV data and ``heart disease risk factors and outcomes'' for the Framingham data. Examples from the real data are provided in a JSON dictionary format, with each feature's name serving as a key for the list of actual values for that feature. This format was used in order to take advantage of the OpenAI API's output format specification. Lastly, the instructions reiterate the model role, prompting the model to generate ``new and diverse samples which
have the correct labels conditioned on the features''.

\begin{table}[]
    \centering
    \begin{tabularx}{\textwidth}{@{} l Y @{}}
    \toprule
        \textbf{Role} & You are a synthetic data generator. Your goal is to produce data which mirrors the given examples in causal structure and feature and label distributions but also produce as diverse samples as possible. I will give you real examples first. \\
        \addlinespace
        \textbf{Context} & Leverage your medical knowledge about hospital admissions and readmission to generate 10 
        realistic but diverse samples [specifically for {Race/Ethnicity} patients]. \\
        \addlinespace
        \textbf{Examples} & \{ \\
        & Age: [27, 68, 13, \dots], \\
        & Sex (Male): [1, 0, 0, \dots], \\
        & \dots , \\
        & CVD: [0, 0, 1, \dots], \\
        & CHD: [0, 1, 0, \dots], \\
        & CHF: [0, 1, 0, \dots] \\
        & \} \\
        \addlinespace
        \textbf{Instructions} & DO NOT COPY THE EXAMPLES but generate realistic but new and diverse samples which
        have the correct labels conditioned on the features. Use the same JSON format as above. \\
    \bottomrule
    \end{tabularx}
    \caption{The prompt template and the text for each of the 4 sections: Role, Context, Examples, and Instructions.}
    \label{tab:prompt}
\end{table}

\section{Temperature Tuning}
\label{app2}
In synthetic data generation, the temperature hyper-parameter controls the randomness and creativity of the synthetic data points generated. For GPT4T, the temperature value ranges from 0 to 2, where 2 represents more varied, random responses and 0 represents more deterministic responses. We conduct experiments to identify the best temperature to produce the most realistic synthetic data samples for each coarse racial group so that it improves the downstream model performance for the racial group. 

We evaluate the temperatures 0.5, 0.9, and 1.2 as they offer a spectrum of conservative to moderately creative, random data points. This allows us to evaluate how much randomness is appropriate when producing realistic synthetic data which is still reflective of the nuances and biases present in the real racial group data. We tested temperatures higher than 1.2, but the data produced consistently violated the JSON formatting schema and some of the data points produced were facially unrealistic. In contrast, for temperatures below 0.5, multiple data points of the same sample were generated, which meant that the data produced was not diverse enough to represent the realistic racial group data.

We perform these temperature experiments on the MIMIC-IV data. We find that mid-range temperatures (0.5 and 0.9) perform better for the critical and ed\_revisit\_3d outcomes across all racial groups, and higher temperatures (1.2) slightly improve the performance for hospitalization prediction outcomes. Critical and readmission outcomes seem to benefit from a more balanced control of randomness, while hospitalization outcomes required capturing more diverse creative data. However, looking at the 0.9 temperature, the AUROC scores are always within 1\% of the scores for the other temperatures even when other temperatures performed better. Thus, since temperature 0.9 strikes a balance between controlled randomness and more creativity, it was the preferred choice for most outcomes as it produced realistic yet diverse synthetic data. We provide the full experiment results of varying temperature with GPT4T generation on the MIMIC-IV data in Table \ref{tab:temp-tests} and we expect similar results on the Framingham dataset.

Through these experiments, we demonstrate that adjusting the temperature hyper-parameter significantly impacts the realism and diversity of the synthetic data produced across different racial groups. Thorough experiments need to be done to identify the best temperature to produce high quality synthetic data for a given task. 

\begin{table}[p]
\begin{adjustwidth}{0.5cm}{0.5cm}
    \centering
    \scriptsize
    \begin{tabularx}{\linewidth}{XXccc} 
    \toprule
     \textbf{Race} & \textbf{Outcome} &  \textbf{Temp = 0.5} & \textbf{Temp = 0.9} & \textbf{Temp = 1.2} \\
    \midrule
       \multirow{3}{*}{Asian}     & critical          & 0.830824 & \textbf{0.846742} & 0.813409 \\
      & ed\_revisit\_3d    & \textbf{0.618036} & 0.611715 & 0.585042 \\
      & hospitalization    & 0.790750 & 0.785137 & \textbf{0.796837} \\
      \cline{1-5}
      \multirow{3}{*}{Black}     & critical          & \textbf{0.845357} & 0.827730 & 0.835718 \\
      & ed\_revisit\_3d    & 0.577016 & \textbf{0.612556} & 0.589905 \\
      & hospitalization    & \textbf{0.794014} & 0.793109 & 0.789897 \\
      \cline{1-5}
      \multirow{3}{*}{Hispanic}  & critical          & \textbf{0.861400} & 0.848560 & 0.849301 \\
      & ed\_revisit\_3d    & 0.664136 & \textbf{0.673516} & 0.634886 \\
      & hospitalization   & 0.785108 & 0.785634 & \textbf{0.790581} \\
    \bottomrule
    \end{tabularx}
    \caption{\textbf{Temperature testing} AUROC results for the GPT-4T model on the MIMIC-IV dataset, across various coarse racial groups and outcome variables. Abbreviations for outcomes are: \textit{critical} (ICU admission or death), \textit{ed\_revisit\_3d} (3-day emergency department readmission), and \textit{hospitalization} (hospital admission).}
    \label{tab:temp-tests}
\end{adjustwidth}
\end{table}

\section{Minority Group Size Sensitivity Analysis}
\label{app:min-size}

We conduct a sensitivity analysis to evaluate whether our augmentation method's performance is affected by the relative size of the minority group compared to the majority group. In the main experiment, the minority group size is 100, so we compare the AUROC results for a lower size of 50 and a higher size of 200. The results are displayed in Table \ref{tab:minority-size} for the MIMIC-IV dataset (the minority groups are too small in Framingham to test any size beyond 100). We find little to no change in performance across any of the methods, and GPT4T continue to outperform across outcomes and racial groups regardless of group size (with the exception of the hospitalization outcome for Asian patients).

\begin{table}[!ht]
\begin{adjustwidth}{-1.5cm}{-1.5cm}
    \centering
    \scriptsize
    \begin{tabularx}{\linewidth}{lllcccccc} % |l|l|l|c|c|c|c||c|c|
        \toprule
        & & & &\multicolumn{3}{c}{Standard Approaches} & \multicolumn{2}{c}{GPT4 Generation} \\ \cmidrule(lr){5-7} \cmidrule(lr){8-9}
        \textbf{Race} & \textbf{Outcome} & \textbf{Size} & \textbf{Baseline} & \textbf{Upweighted} & \textbf{Separate} & \textbf{SMOTE} & \textbf{Group} & \textbf{Generic} \\ \cline{1-9}
\multirow[c]{9}{*}{Asian} & \multirow[c]{3}{*}{critical} & 50 & 0.8739 & 0.8551 & 0.8732 & 0.8555 & \bfseries 0.8778 & 0.8777 \\
 &  & 100 & 0.8744 & 0.8638 & 0.8733 & 0.8661 & 0.8777 & \bfseries 0.8782 \\
 &  & 200 & 0.8752 & 0.8711 & 0.8733 & 0.8680 & \bfseries 0.8790 & 0.8787 \\
\cline{2-9}
 & \multirow[c]{3}{*}{ed\_revisit\_3d} & 50 & 0.5647 & 0.5517 & 0.5634 & 0.5524 & \bfseries 0.5721 & 0.5661 \\
 &  & 100 & 0.5681 & 0.5498 & 0.5624 & 0.5501 & \bfseries 0.5693 & 0.5646 \\
 &  & 200 & 0.5728 & 0.5633 & 0.5637 & 0.5615 & \bfseries 0.5756 & 0.5689 \\
\cline{2-9}
 & \multirow[c]{3}{*}{hospitalization} & 50 & \bfseries 0.8134 & 0.8050 & 0.8133 & 0.8035 & 0.8128 & 0.8132 \\
 &  & 100 & 0.8130 & 0.8100 & 0.8132 & 0.8092 & 0.8132 & \bfseries 0.8139 \\
 &  & 200 & \bfseries 0.8145 & 0.8142 & 0.8131 & 0.8120 & 0.8138 & 0.8142 \\
\cline{1-9}
\multirow[c]{9}{*}{Black} & \multirow[c]{3}{*}{critical} & 50 & 0.8543 & 0.8388 & 0.8534 & 0.8328 & \bfseries 0.8589 & 0.8565 \\
 &  & 100 & 0.8563 & 0.8524 & 0.8551 & 0.8481 & \bfseries 0.8609 & 0.8576 \\
 &  & 200 & 0.8580 & 0.8569 & 0.8521 & 0.8522 & \bfseries 0.8647 & 0.8594 \\
\cline{2-9}
 & \multirow[c]{3}{*}{ed\_revisit\_3d} & 50 & 0.6093 & 0.5709 & 0.6056 & 0.5618 & 0.6316 & \bfseries 0.6321 \\
 &  & 100 & 0.6250 & 0.5925 & 0.6125 & 0.5871 & \bfseries 0.6505 & 0.6430 \\
 &  & 200 & 0.6440 & 0.6211 & 0.6118 & 0.6357 & \bfseries 0.6544 & 0.6543 \\
\cline{2-9}
 & \multirow[c]{3}{*}{hospitalization} & 50 & 0.7994 & 0.7940 & 0.7981 & 0.7928 & \bfseries 0.8022 & 0.8007 \\
 &  & 100 & 0.7990 & 0.7981 & 0.7975 & 0.7966 & \bfseries 0.8024 & 0.8017 \\
 &  & 200 & 0.8026 & 0.8024 & 0.7982 & 0.8007 & 0.8036 & \bfseries 0.8038 \\
\cline{1-9}
\multirow[c]{9}{*}{Hispanic} & \multirow[c]{3}{*}{critical} & 50 & 0.8457 & 0.8374 & 0.8462 & 0.8328 & 0.8536 & \bfseries 0.8545 \\
 &  & 100 & 0.8486 & 0.8419 & 0.8464 & 0.8373 & 0.8534 & \bfseries 0.8554 \\
 &  & 200 & 0.8485 & 0.8483 & 0.8450 & 0.8424 & 0.8546 & \bfseries 0.8554 \\
\cline{2-9}
 & \multirow[c]{3}{*}{ed\_revisit\_3d} & 50 & 0.5854 & 0.5569 & 0.5865 & 0.5472 & 0.6027 & \bfseries 0.6061 \\
 &  & 100 & 0.5858 & 0.5684 & 0.5844 & 0.5674 & 0.6075 & \bfseries 0.6104 \\
 &  & 200 & 0.5969 & 0.5892 & 0.5807 & 0.5826 & 0.6094 & \bfseries 0.6131 \\
\cline{2-9}
 & \multirow[c]{3}{*}{hospitalization} & 50 & 0.7915 & 0.7856 & 0.7903 & 0.7841 & 0.7921 & \bfseries 0.7925 \\
 &  & 100 & 0.7915 & 0.7895 & 0.7893 & 0.7870 & 0.7915 & \bfseries 0.7925 \\
 &  & 200 & 0.7922 & 0.7924 & 0.7878 & 0.7913 & 0.7935 & \bfseries 0.7939 \\ \bottomrule
    \end{tabularx}
    \caption{\textbf{AUROC} scores for each minority racial group, outcome variable, and method for augmenting small groups, across different minority group sizes. Abbreviations for outcomes are critical: Critical event (ICU or death), ed\_revisit\_3d: 3-day emergency department readmission, hospitalization: hospital admission.}
    \label{tab:minority-size}
\end{adjustwidth}
\end{table}

\end{document}

%% file: statement_of_significance.tex
\begin{table}[h]
    \centering
    \footnotesize
    \begin{tabularx}\textwidth{@{} l Y @{}}
    \toprule
        \textbf{Problem or Issue} & Disparities in healthcare prediction model performance can be due to different sized subgroups, prompting health equity concerns. \\
        \textbf{What is Already Known} & Although prior studies have developed methods for tabular health data augmentation with large language models (LLMs), a limitation has been the lack of focus on ensuring augmentation benefits are spread among subgroups. \\
        \textbf{What this Paper Adds} & This study investigates whether LLMs (specifically GPT4-Turbo) can generate group-specific data to augment smaller, underrepresented groups, and improve downstream prediction model performance. The group-specific generation is achieved through the inclusion of group names and group samples in the LLM prompt. \\ 
    \bottomrule
    \end{tabularx}
    \caption{Statement of Significance}
    \label{tab:state_sig}
\end{table}

%% file: arxiv_main.bbl
\begin{thebibliography}{10}

\bibitem{abhari2023mitigating}
Julian Abhari and Ashwin Ashok.
\newblock Mitigating racial biases for machine learning based skin cancer
  detection.
\newblock In {\em Proceedings of the Twenty-fourth International Symposium on
  Theory, Algorithmic Foundations, and Protocol Design for Mobile Networks and
  Mobile Computing}, pages 556--561, 2023.

\bibitem{bastani2021predicting}
Hamsa Bastani.
\newblock Predicting with proxies: Transfer learning in high dimension.
\newblock {\em Management Science}, 67(5):2964--2984, 2021.

\bibitem{beunza2019comparison}
Juan-Jose Beunza, Enrique Puertas, Ester Garc{\'\i}a-Ovejero, Gema Villalba,
  Emilia Condes, Gergana Koleva, Cristian Hurtado, and Manuel~F Landecho.
\newblock Comparison of machine learning algorithms for clinical event
  prediction (risk of coronary heart disease).
\newblock {\em Journal of biomedical informatics}, 97:103257, 2019.

\bibitem{borisov_language_2022}
Vadim Borisov, Kathrin Sessler, Tobias Leemann, Martin Pawelczyk, and Gjergji
  Kasneci.
\newblock Language {Models} are {Realistic} {Tabular} {Data} {Generators}.
\newblock September 2022.

\bibitem{chawla2002smote}
Nitesh~V Chawla, Kevin~W Bowyer, Lawrence~O Hall, and W~Philip Kegelmeyer.
\newblock Smote: synthetic minority over-sampling technique.
\newblock {\em Journal of artificial intelligence research}, 16:321--357, 2002.

\bibitem{chen2018my}
Irene Chen, Fredrik~D Johansson, and David Sontag.
\newblock Why is my classifier discriminatory?
\newblock {\em Advances in neural information processing systems}, 31, 2018.

\bibitem{chen2021ethical}
Irene~Y Chen, Emma Pierson, Sherri Rose, Shalmali Joshi, Kadija Ferryman, and
  Marzyeh Ghassemi.
\newblock Ethical machine learning in healthcare.
\newblock {\em Annual review of biomedical data science}, 4:123--144, 2021.

\bibitem{chen2024benchmarking}
Jiawei Chen, Hongyu Lin, Xianpei Han, and Le~Sun.
\newblock Benchmarking large language models in retrieval-augmented generation.
\newblock In {\em Proceedings of the AAAI Conference on Artificial
  Intelligence}, volume~38, pages 17754--17762, 2024.

\bibitem{chlap_review_2021}
Phillip Chlap, Hang Min, Nym Vandenberg, Jason Dowling, Lois Holloway, and
  Annette Haworth.
\newblock A review of medical image data augmentation techniques for deep
  learning applications.
\newblock {\em Journal of Medical Imaging and Radiation Oncology},
  65(5):545--563, 2021.

\bibitem{chung2024scaling}
Hyung~Won Chung, Le~Hou, Shayne Longpre, Barret Zoph, Yi~Tay, William Fedus,
  Yunxuan Li, Xuezhi Wang, Mostafa Dehghani, Siddhartha Brahma, et~al.
\newblock Scaling instruction-finetuned language models.
\newblock {\em Journal of Machine Learning Research}, 25(70):1--53, 2024.

\bibitem{clemmer_precisedebias_2024}
Colton Clemmer, Junhua Ding, and Yunhe Feng.
\newblock {PreciseDebias}: {An} {Automatic} {Prompt} {Engineering} {Approach}
  for {Generative} {AI} to {Mitigate} {Image} {Demographic} {Biases}.
\newblock In {\em 2024 {IEEE}/{CVF} {Winter} {Conference} on {Applications} of
  {Computer} {Vision} ({WACV})}, pages 8581--8590, Waikoloa, HI, USA, January
  2024. IEEE.

\bibitem{devon2008symptoms}
Holli~A DeVon, Catherine~J Ryan, Amy~L Ochs, and Moshe Shapiro.
\newblock Symptoms across the continuum of acute coronary syndromes:
  differences between women and men.
\newblock {\em American journal of critical care}, 17(1):14--24, 2008.

\bibitem{dondelinger2020joint}
Frank Dondelinger, Sach Mukherjee, and Alzheimer’s Disease~Neuroimaging
  Initiative.
\newblock The joint lasso: high-dimensional regression for group structured
  data.
\newblock {\em Biostatistics}, 21(2):219--235, 2020.

\bibitem{gijsberts2015race}
Crystel~M Gijsberts, Karlijn~A Groenewegen, Imo~E Hoefer, Marinus~JC Eijkemans,
  Folkert~W Asselbergs, Todd~J Anderson, Annie~R Britton, Jacqueline~M Dekker,
  Gunnar Engstr{\"o}m, Greg~W Evans, et~al.
\newblock Race/ethnic differences in the associations of the framingham risk
  factors with carotid imt and cardiovascular events.
\newblock {\em PLoS One}, 10(7):e0132321, 2015.

\bibitem{huang_evaluation_2022}
Jonathan Huang, Galal Galal, Mozziyar Etemadi, and Mahesh Vaidyanathan.
\newblock Evaluation and {Mitigation} of {Racial} {Bias} in {Clinical}
  {Machine} {Learning} {Models}: {Scoping} {Review}.
\newblock {\em JMIR Medical Informatics}, 10(5):e36388, May 2022.

\bibitem{johnson2020mimic}
Alistair Johnson, Lucas Bulgarelli, Tom Pollard, Steven Horng, Leo~Anthony
  Celi, and Roger Mark.
\newblock Mimic-iv.
\newblock {\em PhysioNet. Available online at: https://physionet.
  org/content/mimiciv/1.0/(accessed August 23, 2021)}, pages 49--55, 2020.

\bibitem{juhn_assessing_2022}
Young~J Juhn, Euijung Ryu, Chung-Il Wi, Katherine~S King, Momin Malik, Santiago
  Romero-Brufau, Chunhua Weng, Sunghwan Sohn, Richard~R Sharp, and John~D
  Halamka.
\newblock Assessing socioeconomic bias in machine learning algorithms in health
  care: a case study of the {HOUSES} index.
\newblock {\em Journal of the American Medical Informatics Association},
  29(7):1142--1151, July 2022.

\bibitem{kim2024group}
Jinhee Kim, Taesung Kim, and Jaegul Choo.
\newblock Group-wise prompting for synthetic tabular data generation using
  large language models.
\newblock {\em arXiv preprint arXiv:2404.12404}, 2024.

\bibitem{kleinberg_racial_2022}
Giona Kleinberg, Michael~J. Diaz, Sai Batchu, and Brandon Lucke-Wold.
\newblock Racial underrepresentation in dermatological datasets leads to biased
  machine learning models and inequitable healthcare.
\newblock {\em Journal of biomed research}, 3(1):42, 2022.

\bibitem{ktena2024generative}
Ira Ktena, Olivia Wiles, Isabela Albuquerque, Sylvestre-Alvise Rebuffi, Ryutaro
  Tanno, Abhijit~Guha Roy, Shekoofeh Azizi, Danielle Belgrave, Pushmeet Kohli,
  Taylan Cemgil, et~al.
\newblock Generative models improve fairness of medical classifiers under
  distribution shifts.
\newblock {\em Nature Medicine}, pages 1--8, 2024.

\bibitem{mehrabi2021survey}
Ninareh Mehrabi, Fred Morstatter, Nripsuta Saxena, Kristina Lerman, and Aram
  Galstyan.
\newblock A survey on bias and fairness in machine learning.
\newblock {\em ACM computing surveys (CSUR)}, 54(6):1--35, 2021.

\bibitem{movva_coarse_2023}
Rajiv Movva, Divya Shanmugam, Kaihua Hou, Priya Pathak, John Guttag, Nikhil
  Garg, and Emma Pierson.
\newblock Coarse race data conceals disparities in clinical risk score
  performance, August 2023.

\bibitem{murtaza_synthetic_2023}
Hajra Murtaza, Musharif Ahmed, Naurin~Farooq Khan, Ghulam Murtaza, Saad Zafar,
  and Ambreen Bano.
\newblock Synthetic data generation: {State} of the art in health care domain.
\newblock {\em Computer Science Review}, 48:100546, May 2023.

\bibitem{nazha2019enrollment}
Bassel Nazha, Manoj Mishra, Rebecca Pentz, and Taofeek~K Owonikoko.
\newblock Enrollment of racial minorities in clinical trials: old problem
  assumes new urgency in the age of immunotherapy.
\newblock {\em American Society of Clinical Oncology Educational Book},
  39:3--10, 2019.

\bibitem{park_comparison_2021}
Yoonyoung Park, Jianying Hu, Moninder Singh, Issa Sylla, Irene Dankwa-Mullan,
  Eileen Koski, and Amar~K. Das.
\newblock Comparison of {Methods} to {Reduce} {Bias} {From} {Clinical}
  {Prediction} {Models} of {Postpartum} {Depression}.
\newblock {\em JAMA Network Open}, 4(4):e213909, April 2021.

\bibitem{seedat_curated_2024}
Nabeel Seedat, Nicolas Huynh, Boris van Breugel, and Mihaela van~der Schaar.
\newblock Curated {LLM}: {Synergy} of {LLMs} and {Data} {Curation} for tabular
  augmentation in ultra low-data regimes, February 2024.

\bibitem{singhal2023towards}
Karan Singhal, Tao Tu, Juraj Gottweis, Rory Sayres, Ellery Wulczyn, Le~Hou,
  Kevin Clark, Stephen Pfohl, Heather Cole-Lewis, Darlene Neal, et~al.
\newblock Towards expert-level medical question answering with large language
  models.
\newblock {\em arXiv preprint arXiv:2305.09617}, 2023.

\bibitem{tsao2015cohort}
Connie~W Tsao and Ramachandran~S Vasan.
\newblock Cohort profile: The framingham heart study (fhs): overview of
  milestones in cardiovascular epidemiology.
\newblock {\em International journal of epidemiology}, 44(6):1800--1813, 2015.

\bibitem{ustun2019fairness}
Berk Ustun, Yang Liu, and David Parkes.
\newblock Fairness without harm: Decoupled classifiers with preference
  guarantees.
\newblock In {\em International Conference on Machine Learning}, pages
  6373--6382. PMLR, 2019.

\bibitem{zack_assessing_2024}
Travis Zack, Eric Lehman, Mirac Suzgun, Jorge~A. Rodriguez, Leo~Anthony Celi,
  Judy Gichoya, Dan Jurafsky, Peter Szolovits, David~W. Bates, Raja-Elie~E.
  Abdulnour, Atul~J. Butte, and Emily Alsentzer.
\newblock Assessing the potential of {GPT}-4 to perpetuate racial and gender
  biases in health care: a model evaluation study.
\newblock {\em The Lancet Digital Health}, 6(1):e12--e22, January 2024.

\end{thebibliography}
